%% file: CVPR14w.tex
\documentclass[10pt,twocolumn,letterpaper]{article}

\usepackage{cvpr}
\usepackage{times}
\usepackage{epsfig}
\usepackage{graphicx}
\usepackage{amsmath}
\usepackage{amssymb}
\usepackage{color}
\usepackage{caption}
\usepackage{subcaption}
\usepackage[table]{xcolor}
\usepackage{algorithm,algpseudocode}
\usepackage{rotating}
\usepackage{multirow}
\usepackage{arydshln}
\usepackage{makecell}
\usepackage{booktabs}

\usepackage{booktabs}
\usepackage{pifont}

\usepackage[pagebackref=false,breaklinks=true,letterpaper=true,colorlinks,bookmarks=false]{hyperref}

\def\overfeat{\texttt{OverFeat} }
\def\L2{$L2$}
\def\personA{\textbf{Prof}: }
\def\personB{\textbf{Student}: }
\def\answer{\textbf{Answer}: }
\newcommand{\tickYes}{\checkmark}
\newcommand{\tickNo}{\hspace{1pt}\ding{55}}
\cvprfinalcopy 

\def\cvprPaperID{6} 

\ifcvprfinal\fi
\begin{document}

\title{CNN Features off-the-shelf: an Astounding Baseline for Recognition}

\author{Ali Sharif Razavian \,\,   Hossein Azizpour \,\,   Josephine Sullivan \,\,   Stefan Carlsson\\
CVAP, KTH (Royal Institute of Technology)\\
Stockholm, Sweden\\
{\{\tt\small razavian,azizpour,sullivan,stefanc\}@csc.kth.se}}

\maketitle

\input{Abstract.tex}

\input{Introduction.tex}
\input{Background.tex}
\section{Visual Classification}
Here we go through different tasks related to visual classification in the following subsections.
\label{sec:vis_class}
\input{Method.tex}

\input{Classification.tex}

\input{Detection.tex}
\input{Finegrained.tex}

\input{Attribute.tex}

\input{Implementation.tex}
\input{Retrieval.tex}
\input{Conclusion.tex}
\input{Acknowledgment.tex}

{\small
\bibliographystyle{ieee}
\bibliography{egbib}
}

\end{document}

%% file: Abstract.tex
\begin{abstract}

Recent results indicate that the generic descriptors extracted
from the convolutional neural networks are very powerful.
This paper adds to the mounting evidence that this is
indeed the case. We report on a series of experiments
conducted for different recognition tasks using the
publicly available code and model of the \overfeat network
which was trained to perform object classification on
ILSVRC13. We use features extracted from the \overfeat
 network as a generic image representation to tackle the diverse range of recognition tasks of object image classification,
scene recognition, fine grained recognition, attribute
detection and image retrieval applied to a diverse set of datasets. We selected these
tasks and datasets as they gradually move further away
from the original task and data the \overfeat network was
trained to solve. Astonishingly, we report consistent superior
results compared to the highly tuned state-of-the-art systems in all the visual classification tasks
on various datasets. For instance retrieval it consistently outperforms low memory footprint methods except for sculptures dataset. The results are achieved using a linear
SVM classifier (or $L2$ distance in case of retrieval) applied to a feature representation of size
4096 extracted from a layer in the net. The representations are further 
modified using simple augmentation techniques e.g. jittering. 
The results strongly suggest that features obtained from deep learning with
convolutional nets should be the primary candidate in
most visual recognition tasks.
\end{abstract}

%% file: Introduction.tex
\section{Introduction}

\begin{figure}
  \hspace{-0.8cm}
  \centering
  \begin{subfigure}[b]{0.8\linewidth}
    \centering
    \includegraphics[clip=true, trim=0cm 13cm 1.2cm 0cm, width=1\linewidth]{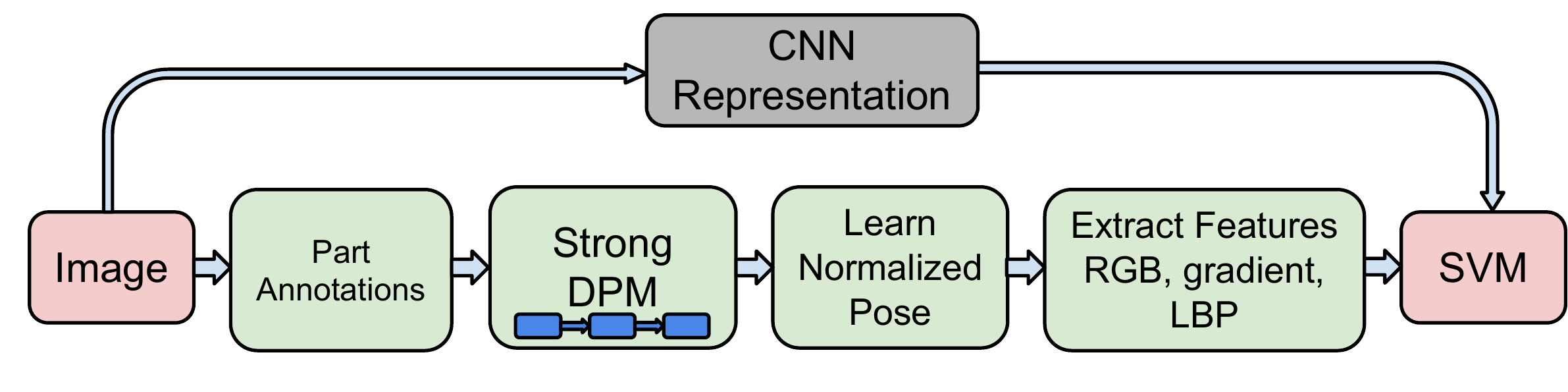}  
      \label{fig:teaser1}
  \end{subfigure}%
  \\
  \vspace{-0.2cm}
  \begin{subfigure}[b]{1\linewidth}
\includegraphics[clip=true, trim=0cm 0cm 0cm 0cm, width=1.05\linewidth]{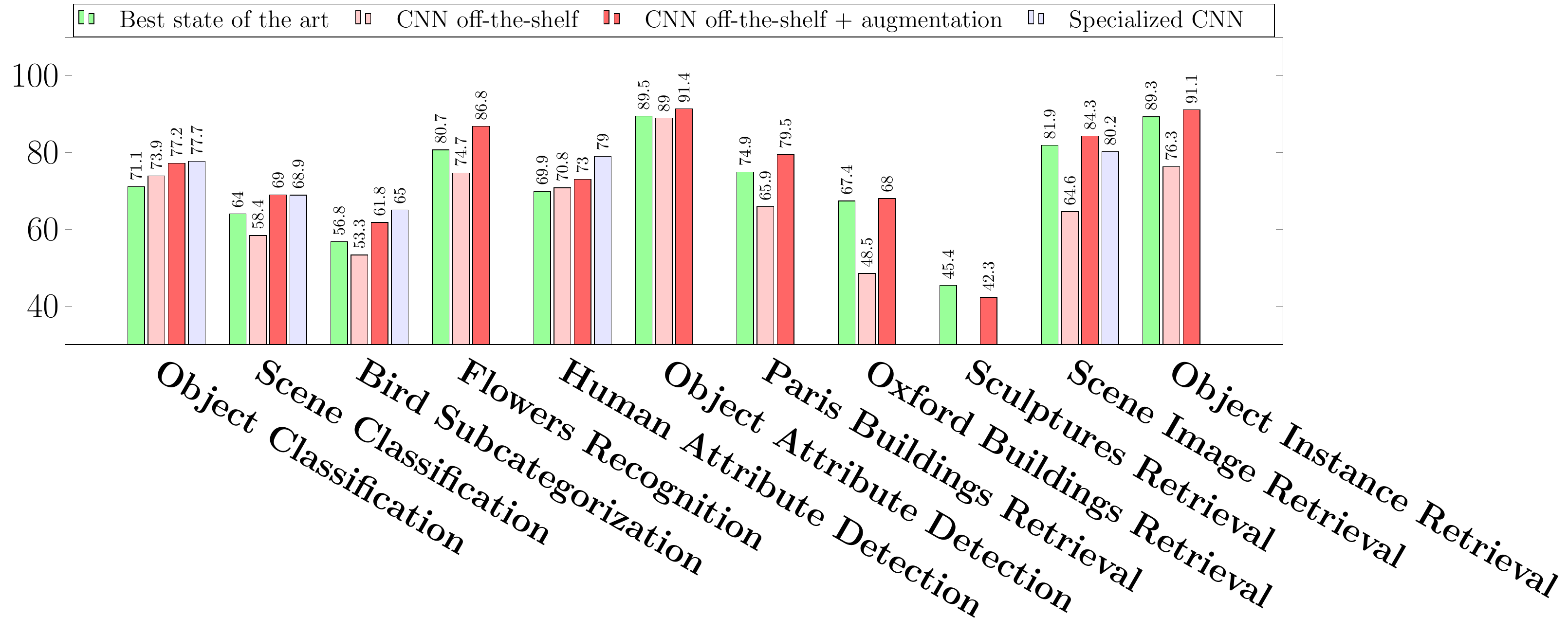}  
  \label{fig:teaser2}
  \end{subfigure}
  \vspace{-0.7cm}
\caption{\footnotesize \textbf{top}) CNN representation replaces pipelines of
            s.o.a methods and achieve better results. e.g. DPD \cite{Zhang13}.\\ \textbf{bottom}) Augmented CNN representation with linear SVM consistently outperforms s.o.a. on multiple tasks. Specialized CNN refers to other works which specifically designed the CNN for their task}
\label{fig:teaser}
  \vspace{-0.4cm}
\end{figure}

\textit{``Deep learning. How well do you think it would work for your
  computer vision problem?"}  Most likely this question has been posed
in your group's coffee room. And in response someone has quoted recent
success stories \cite{Oquab13,Girshick13,Donahue14} and someone else
professed skepticism. You may have left the coffee room slightly
dejected thinking ``Pity I have neither the time, GPU programming
skills nor large amount of labelled data to train my own network to
quickly find out the answer". But when the convolutional neural
network \overfeat \cite{Sermanet13} was recently made publicly
available\footnote{There are other publicly available deep learning implementations such as Alex Krizhevsky's \texttt{ConvNet} and Berkeley's \texttt{Caffe}. Benchmarking these implementations is beyond the scope of this paper.} it allowed for some experimentation. In particular
we wondered now, not whether one could train a deep network
specifically for a given task, but if the features extracted by a deep
network - one carefully trained on the diverse ImageNet database to
perform the specific task of image classification - could be exploited
for a wide variety of vision tasks. 
We now relate our discussions and general findings because
as a computer vision researcher you've probably had the same questions:

\personA First off has anybody else investigated this issue?

\personB Well it turns out Donahue \etal \cite{Donahue14}, Zeiler and
Fergus \cite{Zeiler13} and Oquab \etal \cite{Oquab13} have suggested
that generic features can be extracted from large CNNs and provided
some initial evidence to support this claim. But they have only
considered a small number of visual recognition tasks. It would be fun
to more thoroughly investigate how powerful these CNN features
are. How should we start?


\personA The simplest thing we could try is to extract an image
feature vector from the \overfeat network and combine this with a
simple linear classifier. The feature vector could just be the
responses, with the image as input, from one of the network's final
layers. For which vision tasks do you think this approach would be
effective?

\personB Definitely \textit{image classification}. Several vision
groups have already produced a big jump in performance from the
previous sate-of-the-art methods on Pascal VOC. But maybe fine-tuning
the network was necessary for the jump? I'm going to try it on Pascal
VOC and just to make it a little bit trickier the MIT scene dataset.

\answer \overfeat does a very good job even without fine-tuning
(section \ref{sec:classification} for details).

\personA Okay so that result confirmed previous findings and is
perhaps not so surprising. We asked the \overfeat features to solve a
problem that they were trained to solve. And ImageNet is more-or-less
a superset of Pascal VOC. Though I'm quite impressed by the
indoor scene dataset result. What about a less amenable problem?

\personB I know \textit{fine-grained classification}. Here we want to
distinguish between sub-categories of a category such as the different
species of flowers. Do you think the more generic \overfeat features
have sufficient representational power to pick up the potentially
subtle differences between very similar classes?

\answer It worked great on a standard bird and flower
database. In its most simplistic form it didn't beat the latest best performing methods but it is
a much cleaner solution with ample scope for
improvement. Actually, adopting a set of simple data augmentation techniques (still with linear SVM) beats the best performing methods. Impressive! (Section \ref{sec:fine_grained} for details.)

\personA Next challenge \textit{attribute detection}? Let's
see if the \overfeat features have encoded something about the
semantic properties of people and objects.

\personB Do you think the global CNN features extracted from the
person's bounding box can cope with the articulations and occlusions
present in the H3D dataset. All the best methods do some sort of part
alignment before classification and during training. 

\answer Surprisingly the CNN features on average beat
poselets and a deformable part model for the person attributes
labelled in the H3D dataset. Wow, how did they do that?! They also work
extremely well on the object attribute dataset. Maybe these \overfeat
features do indeed encode attribute information? (Details in
section \ref{sec:attribute}.)

\personA Can we push things even further? Is there a task \overfeat
features should struggle with compared to more established computer
vision systems? Maybe \textit{instance retrieval}. This task drove the
development of the SIFT and VLAD descriptors and the
bag-of-visual-words approach followed swiftly afterwards. Surely these
highly optimized engineered vectors and mid-level features should win
hands down over the generic features?


\personB I don't think CNN features have a chance if we start
comparing to methods that also incorporate 3D geometric
constraints. Let's focus on descriptor performance. Do new school
descriptors beat old school descriptors in the old school descriptors'
backyard?

\answer Very convincing. Ignoring systems that
impose 3D geometry constraints the CNN features are very competitive
on building and holiday datasets (section \ref{sec:retrieval}). Furthermore, 
doing standard instance retrieval feature processing (\ie PCA, whitening, renormalization) 
it shows superior performance compared to low memory footprint methods on all retrieval benchmarks except for
the sculptures dataset.

\personB The take home message from all these results? 

\personA It's all about the features! SIFT and HOG descriptors
produced big performance gains a decade ago and now deep convolutional
features are providing a similar breakthrough for recognition. 
Thus, applying the well-established computer vision procedures
on CNN representations should potentially push the reported results even further.
In any case, if you develop any new algorithm for a recognition task then it \textbf{must} be
compared against the strong baseline of \textit{generic deep features}
+ \textit{simple classifier}. 

%% file: Background.tex
\section{Background and Outline}
In this work we use the publicly available trained CNN called
\overfeat \cite{Sermanet13}. The structure of this network follows
that of Krizhevsky \etal \cite{Krizhevsky12}. The convolutional layers
each contain 96 to 1024 kernels of size 3$\times$3 to
7$\times$7. Half-wave rectification is used as the nonlinear
activation function. Max pooling kernels of size 3$\times$3 and
5$\times$5 are used at different layers to build robustness to
intra-class deformations. We used the ``large" version of the
\overfeat network. It takes as input color images of size
221$\times$221. Please consult \cite{Sermanet13} and
\cite{Krizhevsky12} for further details. 

\overfeat was trained for the image classification task of ImageNet ILSVRC 2013 \cite{ilsvrc-2013} and obtained very competitive results for the classification task of the 2013 challenge and won the localization task. ILSVRC13
contains 1.2 million images which are hand labelled with the
presence/absence of 1000 categories. The images are mostly centered
and the dataset is considered less challenging in terms of clutter and
occlusion than other object recognition datasets such as PASCAL VOC
\cite{voc-2012}.

We report results on a series of experiments we conducted on different
recognition tasks. The tasks and datasets were selected such that they
gradually move further away from the task the \overfeat network was
trained to perform. We have two sections for visual classification (Sec. \ref{sec:vis_class}) and visual instance retrieval (Sec. \ref{sec:retrieval}) where we review different
tasks and datasets and report the final results. The crucial thing to remember is
that the CNN features used are trained only using ImageNet data though
the simple classifiers are trained using images specific to the task's
dataset.\\ Finally, we have to point out that, given enough
computational resources, optimizing the CNN features for specific
tasks/datasets would probably boost the performance of the simplistic system even
further \cite{Oquab13,Girshick13,Zhang14,Toshev14,Taigman14}.

%% file: Method.tex
\subsection{Method}
\label{sec:method}
For all the experiments,
unless stated otherwise, we use the first fully connected layer (layer 22) of the network as our feature vector. Note the
max-pooling and rectification operations are each considered as a
separate layer in \overfeat which differs from Alex Krizhevsky's ConvNet numbering. For all the experiments we resize the whole image (or cropped sub-window) to 221$\times$221. This gives a vector of 4096 dimensions. We have two settings:
\begin{itemize}
\item The feature vector is
further \L2 normalized to unit length for all the experiments. We use
the 4096 dimensional feature vector in combination with a Support
Vector Machine (SVM) to solve different classification tasks (CNN-SVM). 
\item We further augment the training set by adding cropped and rotated samples and doing component-wise power transform and report separate results (CNNaug+SVM).
\end{itemize}
For the classification scenarios where the labels are not mutually
exclusive (\eg VOC Object Classification or UIUC Object attributes)
we use a one-against-all strategy, in the rest of experiments we use
one-against-one linear SVMs with voting. For all the experiments we
use a linear SVM found from eq.\ref{eq:svm}, where we have training data $\{(\mathbf{x}_i, y_i)\}$.
\small
\begin{align}
\label{eq:svm}
\underset{\mathbf{w}}{\text{minimize}}\,\,\frac{1}{2}\|\mathbf{w}\|^2+C\sum_i\text{max}(1-y_i\mathbf{w}^T\mathbf{x}_i,0)
\end{align}
\normalsize
Further information can be found in the implementation details at section \ref{sec:imp}.

%% file: Classification.tex
\subsection{Image Classification}
\label{sec:classification}
To begin, we adopt the CNN representation to tackle the problem of image classification of objects and scenes. The system should assign
(potentially multiple) semantic labels to an image. Remember in
contrast to object detection, object image classification requires no
localization of the objects. The CNN representation has been optimized for the object image classification task of ILSVRC. Therefore, in this
experiment the representation is more aligned with the
final task than the rest of experiments. However, we have chosen
two different image classification datasets, objects and indoor scenes,
whose image distributions differ from that of ILSVRC dataset.
\subsubsection{Datasets}
\input{Classification_ds.tex}
\input{Classification_r.tex}

%% file: Classification_ds.tex
\begin{table*}
\vspace{-.35cm}
\scriptsize
\centering
\tabcolsep=0.1cm
\begin{tabular}{ l c c c c c c c c c c c c c c c c c c c c c }
  \toprule
     &aero&bike&bird&boat&bottle&bus&car&cat&chair&cow&table&
     dog&horse&mbike&person&plant&sheep&sofa&train&tv&mAP \\ 
     \cmidrule{2-22}
     \rowcolor[gray]{0.8}
	    GHM\cite{Chen12}&76.7&74.7&53.8&72.1&40.4&71.7&83.6&66.5&52.5&57.5
	    &62.8&51.1&81.4&71.5&86.5&36.4&55.3&60.6&80.6&57.8&64.7\\
		AGS\cite{Dong13}&82.2&83.0&58.4&76.1&\textbf{56.4}&\textbf{77.5}&
		\textbf{88.8}&69.1&\textbf{62.2}&61.8&64.2&51.3&\textbf{85.4}&
		\textbf{80.2}&91.1&48.1&61.7&\textbf{67.7}&86.3&70.9&71.1\\
     \rowcolor[gray]{0.8}
		NUS\cite{Song11}&82.5&79.6&64.8&73.4&54.2&75.0&77.5&79.2&46.2&62.7
		&41.4&74.6&85.0&76.8&91.1&53.9&61.0&67.5&83.6&70.6&70.5\\
		\midrule
		CNN-SVM&88.5&81.0&83.5&82.0&42.0&72.5&85.3&81.6&
		59.9&58.5&66.5&77.8&81.8&78.8&90.2&54.8&71.1&62.6&87.2&71.8&73.9\\
     \rowcolor[gray]{0.8}
		CNNaug-SVM&\textbf{90.1}&\textbf{84.4}&\textbf{86.5}&\textbf{84.1}&48.4&73.4&86.7&\textbf{85.4}&
		61.3&\textbf{67.6}&\textbf{69.6}&\textbf{84.0}&\textbf{85.4}&80.0&\textbf{92.0}&\textbf{56.9}&\textbf{76.7}&67.3&\textbf{89.1}&\textbf{74.9}&\textbf{77.2}\\
\bottomrule
\end{tabular}
\vspace{-.2cm}
\caption{{\footnotesize \textbf{Pascal VOC 2007 Image Classification Results} compared to other methods
which also use training data outside VOC. The CNN representation is not
tuned for the Pascal VOC dataset. However, GHM \cite{Chen12} learns from VOC a joint representation of bag-of-visual-words and contextual information. AGS \cite{Dong13} learns a second layer of representation by clustering the VOC data into subcategories. NUS \cite{Song11} trains a codebook for the SIFT, HOG and LBP descriptors from the VOC dataset. Oquab \etal ~\cite{Oquab13} fixes all the layers trained on ImageNet then it adds and optimizes two fully connected layers on the VOC dataset and achieves better results (\textbf{77.7}) indicating the potential to boost the performance by further adaptation of the representation to the target task/dataset.}}
\vspace{-0.5cm}
\label{tab:classification_2007}
\end{table*}
We use two challenging recognition datasets,
Namely, Pascal VOC 2007 for object image classification
\cite{voc-2012} and the MIT-67 indoor scenes \cite{Quattoni09} for
scene recognition.\\
\textbf{Pascal VOC.}
Pascal VOC 2007 \cite{voc-2012} contains $\sim$10000 images of 20 classes including
animals, handmade and natural objects. The objects are not centered and in general the appearance of objects in VOC is
perceived to be more challenging than ILSVRC. Pascal VOC images
come with bounding box annotation which are \textit{not} used in our experiments.\\
\textbf{MIT-67 indoor scenes.} The MIT scenes dataset has
15620 images of 67 indoor scene classes. 
The dataset consists of different types of stores (\eg bakery, grocery)
residential rooms (\eg nursery room, bedroom), public spaces (\eg
inside bus, library, prison cell), leisure places (\eg buffet,
fastfood, bar, movietheater) and working places (\eg office, operating
room, tv studio). The similarity of the objects present in different
indoor scenes makes MIT indoor an especially difficult dataset
compared to outdoor scene datasets. 

%% file: Classification_r.tex
\subsubsection{Results of PASCAL VOC Object Classification}
\begin{figure}
  \centering
  \begin{subfigure}[b]{0.18\textwidth}
  \includegraphics[clip=true, trim=2cm 6cm 2.5cm 6.5cm, width=1\textwidth]{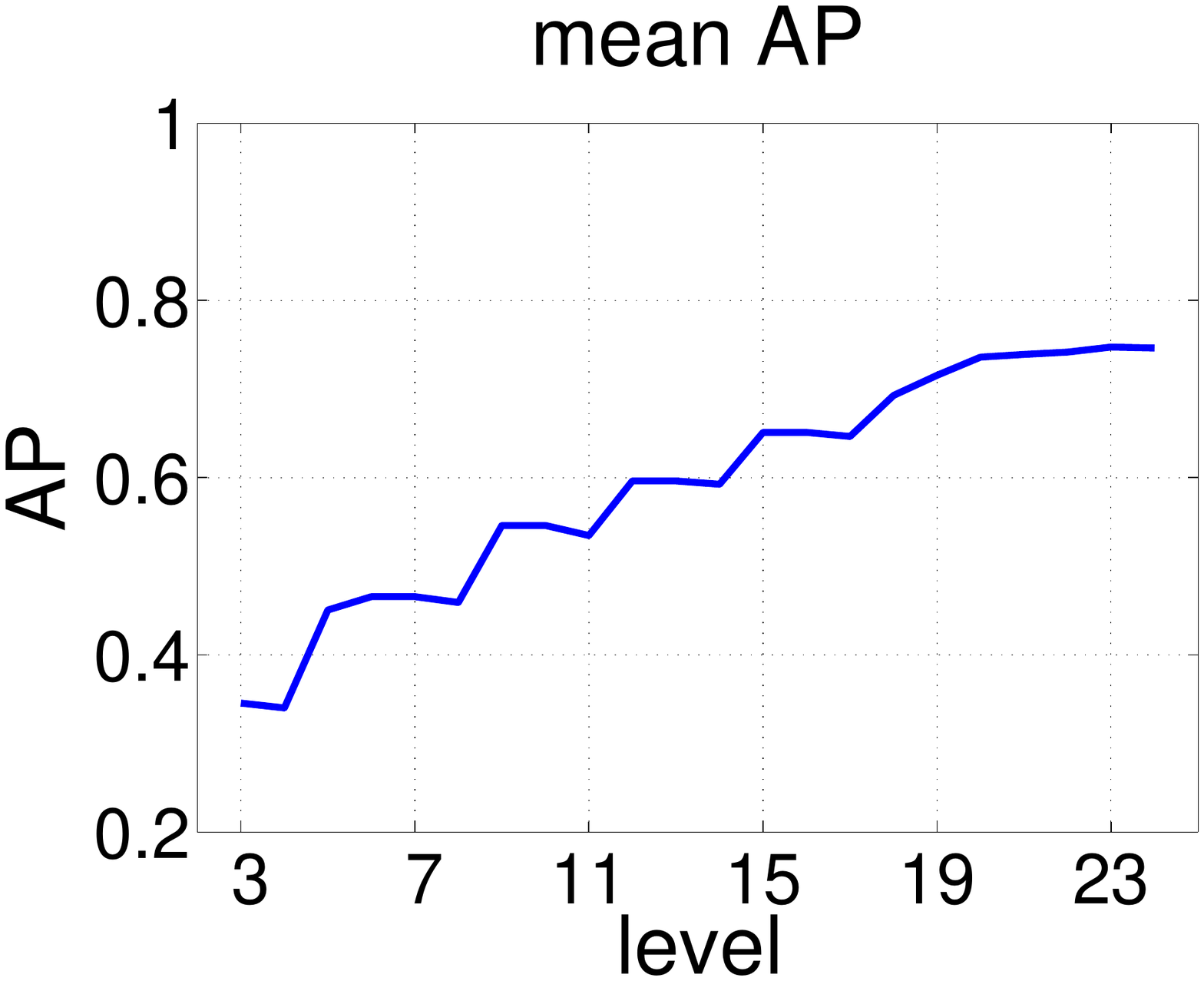}  
  \caption{}
  \label{fig:meanAPrev}
  \end{subfigure} 
  \,\,\,\,\,
  \begin{subfigure}[b]{0.27\textwidth}
  \includegraphics[clip=true, trim=10cm 3.5cm 9cm 0cm, width=1.1\textwidth]
	{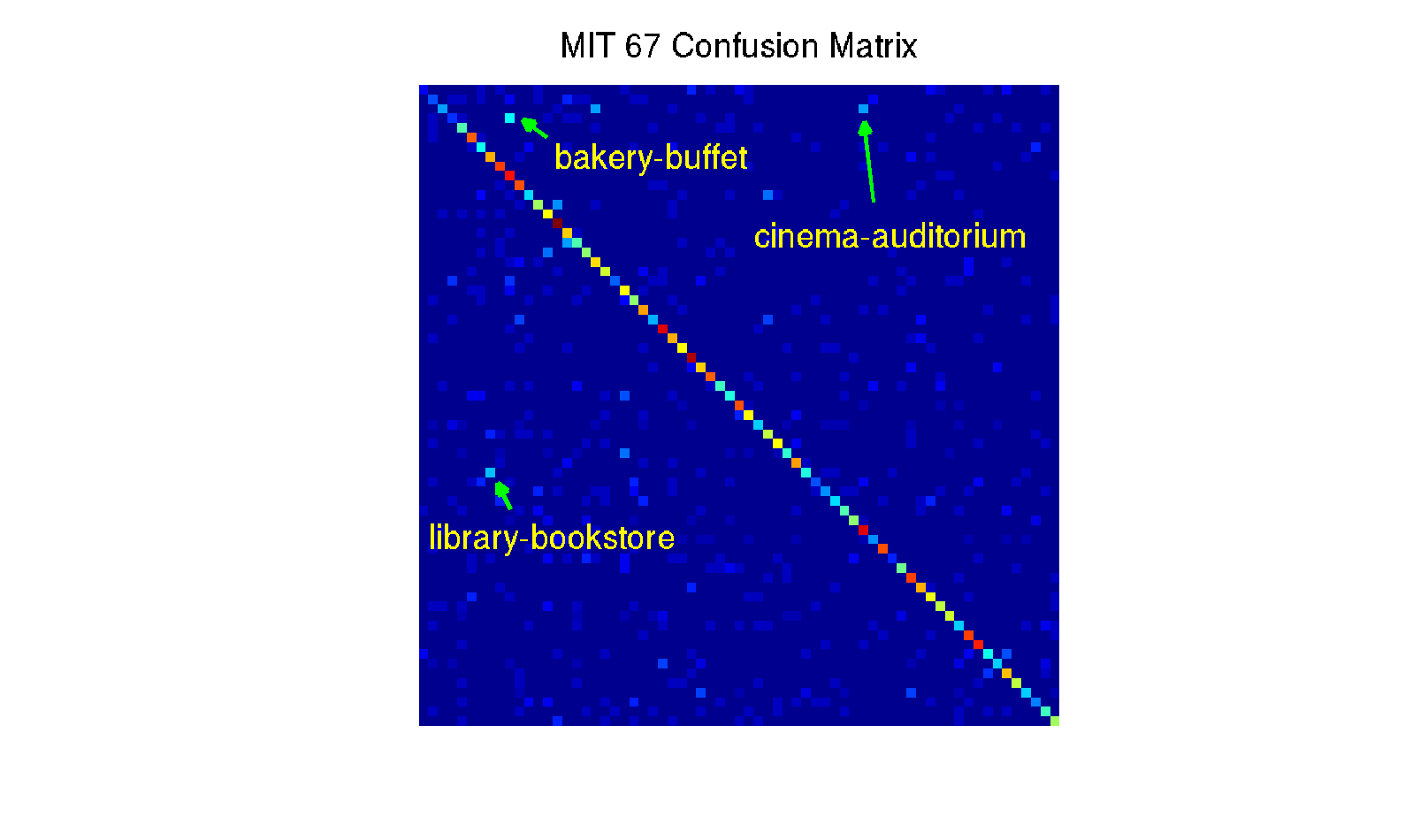}
	\caption{}
	\label{fig:mit_conf}
  \end{subfigure}
  \caption{\footnotesize \textbf{a)} Evolution of the mean image classification AP over PASCAL
    VOC 2007 classes as we use a deeper representation from
    the \overfeat CNN trained on the ILSVRC dataset. \overfeat
	considers convolution, max pooling, nonlinear activations,
	etc. as separate layers. The re-occurring decreases in the
	plot is of the activation function layer which loses
	information by half rectifying the signal. \textbf{b)}
	Confusion matrix for the MIT-67 indoor dataset. Some of the
	off-diagonal confused classes have been annotated, these
	particular cases could be hard even for a human to distinguish.}
    \vspace{-0.3cm}
\end{figure}

Table \ref{tab:classification_2007} shows
the results of the \overfeat CNN representation for object image
classification. The performance is measured using average precision
(AP) criterion of VOC 2007 \cite{voc-2012}. Since the original
representation has been trained for the same task (on ILSVRC) we
expect the results to be relatively high. We compare the results only
with those methods which have used training data outside the standard Pascal VOC 2007 dataset. We can see that the method outperforms all the previous efforts by a significant margin in mean average precision (mAP). Furthermore, it has superior average precision on 10 out of 20 classes. It is worth mentioning the baselines in Table \ref{tab:classification_2007} use sophisticated matching systems. The same observation has been recently made in another work \cite{Oquab13}.
\paragraph{Different layers.} Intuitively one could reason that the
learnt weights for the deeper layers could become more specific to the
images of the training dataset and the task it is trained for. Thus,
one could imagine the optimal representation for each problem lies at
an intermediate level of the network. To further study this, we
trained a linear SVM for all classes using the output of each network
layer. The result is shown in Figure \ref{fig:meanAPrev}. Except for
the fully connected last 2 layers the performance increases. We
observed the same trend in the individual class plots. The subtle
drops in the mid layers (\eg 4, 8, etc.) is due to the ``ReLU" layer
which half-rectifies the signals. Although this will help the
non-linearity of the trained model in the CNN, it does not help if
immediately used for classification.

\subsubsection{Results of MIT 67 Scene Classification}
Table \ref{tab:mit_67} shows the results of
  different methods on the MIT indoor dataset. The performance is
  measured by the average classification accuracy of different classes
  (mean of the confusion matrix diagonal). Using a CNN off-the-shelf
  representation with linear SVMs training significantly outperforms
  a majority of the baselines. The non-CNN baselines benefit from a
  broad range of sophisticated designs. 
  confusion matrix of the CNN-SVM classifier on the 67 MIT classes. It
  has a strong diagonal. The few relatively bright off-diagonal points
  are annotated with their ground truth and estimated labels. One can
  see that in these examples the two labels could be challenging even
  for a human to distinguish between, especially for close-up views of
  the scenes.
\begin{table}[!t]
\footnotesize
\centering
\tabcolsep=0.15cm
\mbox{}\\
    \begin{tabular}{lc}        
     \toprule                  
     Method&mean Accuracy \\ 
     \midrule
     \rowcolor[gray]{0.8}
	    ROI + Gist\cite{Quattoni09}&26.1\\
	    DPM\cite{Pandey11}&30.4\\
     \rowcolor[gray]{0.8}
	    Object Bank\cite{Li10}&37.6\\
     	RBow\cite{Parizi12}&37.9\\
     \rowcolor[gray]{0.8}
	    BoP\cite{Juneja13}&46.1\\
     	miSVM\cite{Li13}&46.4\\
     \rowcolor[gray]{0.8}
	    D-Parts\cite{Sun13}&51.4\\
     	IFV\cite{Juneja13}&60.8\\
     \rowcolor[gray]{0.8}
     	MLrep\cite{Doersch13}&64.0\\
     	\midrule
		CNN-SVM&58.4\\
     \rowcolor[gray]{0.8}
		CNNaug-SVM&\textbf{69.0}\\
		CNN(AlexConvNet)+multiscale pooling \cite{Gong14}&68.9\\
         \bottomrule
\end{tabular}
\caption{{\footnotesize \textbf{MIT-67 indoor scenes dataset}.  The MLrep \cite{Doersch13} has a fine tuned pipeline which takes weeks to select and train various part
detectors. Furthermore, Improved Fisher Vector (IFV) representation has dimensionality larger than 200K. \cite{Gong14} has very recently tuned a multi-scale orderless pooling of CNN features (off-the-shelf) suitable for certain tasks. With this simple modification they achieved significant average classification accuracy of \textbf{68.88}.}}
\label{tab:mit_67}
\vspace{-0.3cm}
\end{table}

%% file: Detection.tex
\vspace{-0.2cm}
\subsection{Object Detection}
Unfortunately, we have not conducted any experiments for using CNN
off-the-shelf features for the task of object detection. But it is
worth mentioning that Girshick \etal \cite{Girshick13} have reported
remarkable numbers on PASCAL VOC 2007 using off-the-shelf features
from \texttt{Caffe} code. We repeat their relevant results here. Using
off-the-shelf features they achieve a mAP of \textbf{46.2} which
already outperforms state of the art by about 10\%. This adds to our
evidences of how powerful the CNN features off-the-shelf are for visual recognition tasks. \\
Finally, by further fine-tuning the representation for PASCAL VOC 2007 dataset (not off-the-shelf anymore) they achieve impressive results of \textbf{53.1}.

%% file: Finegrained.tex
\subsection{Fine grained Recognition}
\label{sec:fine_grained}
Fine grained recognition has recently become popular due to 
its huge potential for both commercial and
cataloging applications. Fine grained recognition is specially
interesting because it involves recognizing subclasses of the same
object class such as different bird species, dog breeds, flower types,
etc. The advent of many new datasets with fine-grained annotations
such as Oxford flowers \cite{Nilsback08}, Caltech bird species
\cite{Wah11}, dog breeds \cite{ilsvrc-2013}, cooking activities
\cite{Rohrbach12}, cats and dogs \cite{parkhi12} has helped the field
develop quickly. The subtlety of differences across different
subordinate classes (as opposed to different categories) requires a
fine-detailed representation. This characteristic makes fine-grained
recognition a good test of whether a generic representation can
capture these subtle details.
\vspace{-0.25cm}
\subsubsection{Datasets}
\input{Finegrained_ds.tex}
\vspace{-0.3cm}
\subsubsection{Results}
\input{Finegrained_r.tex}

%% file: Finegrained_ds.tex
We evaluate CNN features on two fine-grained recognition datasets CUB 200-2011 and 102 Flowers. \\
\textbf{Caltech-UCSD Birds (CUB) 200-2011} dataset \cite{Wah11}
is chosen since many recent methods have reported performance on it. 
It contains 11,788 images of 200 bird subordinates. 5994 images
are used for training and 5794 for evaluation. Many of the species in
the dataset exhibit extremely subtle differences which are sometimes
even hard for humans to distinguish.
Multiple levels of annotation are available for this dataset - bird
bounding boxes, 15 part landmarks, 312 binary attributes and boundary
segmentation. The majority of the methods applied use the bounding box
and part landmarks for training. In this work
we only use the bounding box annotation during training and testing.\\
\textbf{Oxford 102 flowers} dataset \cite{Nilsback08} contains 102 categories. 
Each category contains 40 to 258 of images. The flowers appear at
different scales, pose and lighting conditions. Furthermore, the dataset
provides segmentation for all the images.


%% file: Finegrained_r.tex
\vspace{-0.1cm}
Table \ref{tab:cub_2011} reports the results of the CNN-SVM compared
to the top performing baselines on the CUB 200-2011 dataset. The first
two entries of the table represent the methods which only use bounding
box annotations. The rest of baselines use part annotations for
training and sometimes for evaluation as well. 

Table \ref{tab:flowers_102} shows the performance of CNN-SVM and other
baselines on the flowers dataset. All methods, bar the CNN-SVM, use
the segmentation of the flower from the background. It can be seen
that CNN-SVM outperforms all basic representations and their multiple
kernel combination even without using segmentation.

\begin{table}[!t]
\footnotesize
\centering
\tabcolsep=0.25cm
\mbox{}\\
\begin{tabular}{l c c}
     \toprule
     Method& Part info & mean Accuracy \\
     \midrule 
     \rowcolor[gray]{0.8}
	    Sift+Color+SVM\cite{Wah11} & \tickNo &17.3\\
	    Pose pooling kernel\cite{Zhang12}& \tickYes &28.2\\
     \rowcolor[gray]{0.8}
     	RF\cite{Yao11}& \tickYes &19.2\\
	    DPD\cite{Zhang13}& \tickYes&51.0\\
     \rowcolor[gray]{0.8}
     	Poof\cite{Berg13}& \tickYes&56.8\\
	\midrule
		CNN-SVM& \tickNo & 53.3\\
     \rowcolor[gray]{0.8}
		CNNaug-SVM& \tickNo & \textbf{61.8}\\
	    DPD+CNN(DeCaf)+LogReg\cite{Donahue14}& \tickYes&\textbf{65.0}\\
\bottomrule
\end{tabular}
\caption{{\footnotesize \textbf{Results on CUB 200-2011 Bird dataset.} The 
table distinguishes between methods which use part annotations for
training and sometimes for evaluation as well and those that do not. \cite{Donahue14} generates a pose-normalized CNN representation using DPD~\cite{Zhang13} detectors which significantly boosts the results to \textbf{64.96}.}}
\label{tab:cub_2011}
\end{table}

\begin{table}[!t]
\footnotesize
\centering
\tabcolsep=0.15cm
\mbox{}\\
     \begin{tabular}{l c}        
     \toprule
     Method&mean Accuracy \\ 
     \midrule
     \rowcolor[gray]{0.8}
	    HSV \cite{Nilsback08}&43.0\\
	    SIFT internal \cite{Nilsback08}&55.1\\
     \rowcolor[gray]{0.8}
	    SIFT boundary \cite{Nilsback08}&32.0\\
	    HOG \cite{Nilsback08}&49.6\\
     \rowcolor[gray]{0.8}
	    HSV+SIFTi+SIFTb+HOG(MKL) \cite{Nilsback08}&72.8\\
	    BOW(4000) \cite{Fernando14}&65.5\\
     \rowcolor[gray]{0.8}
	    SPM(4000) \cite{Fernando14}&67.4\\
	    FLH(100) \cite{Fernando14}&72.7\\
     \rowcolor[gray]{0.8}
	    BiCos seg \cite{Chai11}&79.4\\
	    Dense HOG+Coding+Pooling\cite{Angelova13} w/o seg& 76.7\\
     \rowcolor[gray]{0.8}
	    Seg+Dense HOG+Coding+Pooling\cite{Angelova13}& 80.7\\
	    \midrule
	    CNN-SVM w/o seg&74.7\\
     \rowcolor[gray]{0.8}
	    CNNaug-SVM w/o seg&\textbf{86.8}\\
            \bottomrule
\end{tabular}
\caption{{\footnotesize \textbf{Results on the Oxford 102 Flowers dataset.} All the methods use segmentation to subtract the flowers from background unless stated otherwise.}}
\label{tab:flowers_102}
\vspace{-0.5cm}
\end{table}

%% file: Attribute.tex
\subsection{Attribute Detection}
\label{sec:attribute}
An attribute within the context of computer vision is defined as some
semantic or abstract quality which different instances/categories
share.
\subsubsection{Datasets}
\input{Attribute_ds.tex}
\vspace{-0.6cm}
\subsubsection{Results}
\input{Attribute_r.tex}

%% file: Attribute_ds.tex
We use two datasets for
attribute detection. The first dataset is the UIUC 64 object
attributes dataset \cite{Farhadi09}. There are 3 categories of attributes in
this dataset: shape (\eg is 2D boxy), part (\eg has head) or material
(\eg is furry). The second dataset is the H3D dataset \cite{Bourdev11}
which defines 9 attributes for a subset of the person images from Pascal VOC 2007. The attributes range from ``has glasses" to ``is male".

%% file: Attribute_r.tex
Table \ref{tab:objatt} compares CNN features performance to
state-of-the-art. Results are reported for both across
and within categories attribute detection (refer to \cite{Farhadi09}
for details).

Table \ref{tab:h3datt} reports the results of the detection of 9 human
attributes on the H3D dataset including poselets and
DPD \cite{Zhang13}. Both poselets and DPD use part-level annotations
during training while for the CNN we only extract one feature from
the bounding box around the person. The CNN representation performs as
well as DPD and significantly outperforms poselets.
\begin{table}[!t]
\footnotesize
\centering
\tabcolsep=0.15cm
\mbox{}\\
\begin{tabular}{l c c c}
     \toprule      
     Method & within categ. & across categ.&mAUC\\ 
     \midrule
     \rowcolor[gray]{0.8}
		Farhadi \etal \cite{Farhadi09}&83.4&-&73.0\\
		Latent Model\cite{Wang10}&62.2&79.9&-\\
     \rowcolor[gray]{0.8}
		Sparse Representation\cite{Tsagkatakis10}&89.6&\textbf{90.2}&-\\
		att. based classification\cite{Lampert14}&-&-&73.7\\
     \rowcolor[gray]{0.8}
     \midrule
		CNN-SVM&91.7&82.2&89.0\\
		CNNaug-SVM&\textbf{93.7}&84.9&\textbf{91.5}\\
         \bottomrule
\end{tabular}
\caption{{\footnotesize \textbf{UIUC 64 object attribute dataset results}. Compared to
     other existing methods the CNN features perform very favorably.}}
\label{tab:objatt}
\end{table}

\begin{table}[!t]
\scriptsize
\centering
\tabcolsep=0.05cm
\mbox{}\\
\begin{tabular}{lcccccccccc}
\toprule
     Method&male&lg hair&glasses& \, hat \,  &tshirt&lg
     slvs&shorts&jeans&lg pants&mAP \\ 
     \midrule
          \rowcolor[gray]{0.8}
     Freq\cite{Bourdev11}&59.3&30.0&22.0&16.6&23.5&49.0&17.9&33.8&74.7&36.3\\
     SPM\cite{Bourdev11}&68.1&40.0&25.9&35.3&30.6&58.0&31.4&39.5&84.3&45.9\\
          \rowcolor[gray]{0.8}
     Poselets\cite{Bourdev11}&82.4&\textbf{72.5}&\textbf{55.6}&60.1&51.2&74.2&45.5&54.7&90.3&65.2\\
     DPD\cite{Zhang13}&83.7&70.0&38.1&\textbf{73.4}&49.8&78.1&64.1&\textbf{78.1}&93.5&69.9\\
          \rowcolor[gray]{0.8}
          \midrule
	    CNN-SVM&83.0&67.6&39.7&66.8&52.6&82.2&78.2&71.7&95.2&70.8\\
	    CNNaug-SVM&\textbf{84.8}&71.0&42.5&66.9&\textbf{57.7}&\textbf{84.0}&\textbf{79.1}&75.7&\textbf{95.3}&\textbf{73.0}\\
\bottomrule
\end{tabular}
\caption{{\footnotesize\textbf{ H3D Human Attributes dataset results.} A CNN representation
     is extracted from the bounding box surrounding the person. All the
     other methods require the part annotations during training. The
     first row shows the performance of a random classifier. The work of Zhang \etal \cite{Zhang14} has adapted the CNN architecture specifically for the task of attribute detection and achieved the impressive performance of \textbf{78.98} in mAP. This further highlights the importance of adapting the CNN architecture for different tasks given enough computational resources.}}
\label{tab:h3datt}
\vspace{-0.52cm}
\end{table}

%% file: Implementation.tex
\subsection{Implementation Details}
\label{sec:imp}
\vspace{-0.1cm}
We have used precomputed linear kernels with libsvm for the
CNN-SVM experiments and liblinear for the CNNaug-SVM with the primal
solver (\#samples $\gg$ \#dim). Data augmentation is done by making 16
representations for each sample (original image, 5 crops, 2 rotation
and their mirrors). The cropping is done such that the subwindow
contains 4/9 of the original image area from the 4 corners and the
center. We noted the following phenomenon for all
datasets. At the test time, when we have multiple
representations for a test image, taking the sum over all the
responses works outperforms taking the max response. In CNNaug-SVM we
use \textit{signed} component-wise power transform by raising each dimension to
the power of 2. For the datasets which with bounding box (\ie birds,
H3D) we enlarged the bounding box by 150\% to include some context. In
the early stages of our experiments we noticed that using one-vs-one
approach works better than structured SVM for multi-class
learning. Finally, we noticed that using the imagemagick library for image resizing has slight adverse effects compared to matlab imresize function. The cross-validated SVM parameter ($C$) used for different datasets are as follows. VOC2007:\textbf{0.2}, MIT67:\textbf{2} , Birds:\textbf{2}, Flowers:\textbf{2}, H3D:\textbf{0.2} UIUCatt:\textbf{0.2}.\footnote{The details of our system including extracted features, scripts and updated tables can be found at our project webpage: http://www.csc.kth.se/cvap/cvg/DL/ots/}

%% file: Retrieval.tex
\section{Visual Instance Retrieval}
\label{sec:retrieval}
In this section we compare the CNN representation to the current
state-of-the-art retrieval pipelines including VLAD\cite{Arandjelovic13,Zhao13}, BoW, IFV\cite{perronnin10}, Hamming Embedding\cite{Jain11} and BoB\cite{Arandjelovic11}.
Unlike the CNN representation, all the above methods use dictionaries
trained on similar or same dataset as they are tested on. For a fair comparison between the methods, we only report results on representations with relevant order of dimensions and exclude post-processing methods like spatial re-ranking and query expansion.
\subsection{Datasets}
\input{Retrieval_ds.tex}

\subsection{Method}
\input{Retrieval_m.tex}

\subsection{Results}
\input{Retrieval_r.tex}

%% file: Retrieval_ds.tex
We report retrieval results on five common datasets in the area as follows:

\textbf{Oxford5k buildings}\cite{Philbin07}
This is a collection of 5063 reference photos gathered from flickr, and 55 queries of different buildings. From an
architectural standpoint the buildings in Oxford5k are very similar. 
Therefore it is a challenging benchmark for generic features such as CNN.

\textbf{Paris6k buildings}\cite{Philbin08}
Similar to the Oxford5k, this collection has 55 queries images of
buildings and monuments from Paris and 6412 reference photos.
The landmarks in Paris6k have more diversity than those in Oxford5k.

\textbf{Sculptures6k}\cite{Arandjelovic11}
This dataset brings the challenge of smooth and texture-less item
retrieval.  It has 70 query images and contains 6340 reference images which is halved to train/test subsets. 
 The results on this dataset highlights the extent to which CNN features are able to encode shape.


\textbf{Holidays dataset}\cite{jegou08} This dataset contains 1491 images of which 500 are queries. It contains images of different scenes, items and monuments.
Unlike the first three datasets, it exhibits a diverse
set of images. For the above datasets we reported mAP as the measurement metric.

\textbf{UKbench}\cite{nister06}
A dataset of images of 2250 items each from four different viewpoints. The UKbench provides a good benchmark for viewpoint changes. We
reported recall at top four as the performance over UKBench.

%% file: Retrieval_m.tex
Similar to the previous tasks we use the \L2 normalized output of the
first fully connected layer as representation.\\
\textbf{Spatial search.} The items of interest can appear at different
locations and scales in the test and reference images making some form
of spatial search necessary. Our crude search has the following
form. For each image we extract multiple sub-patches of different
sizes at different locations. Let $h$ (the number of levels) represent
the number of different sized patches we extract. At level $i$, $1\leq
i \leq h$, we extract $i^2$ overlapping sub-patches of the same size
whose union covers the whole image. For each extracted sub-patch we
compute its CNN representation. The distance between a query sub-patch and a reference image is defined as the minimum \L2 distance between the query sub-patch and respective reference sub-patches. 
Then, the distance between the reference and the query image is set to 
the average distance of each query sub-patch to the reference image.
In contrast to visual classification pipelines, we extract features from the smallest
square containing the region of interest (as opposed to resizing). In the reset of the text, $h_r$ denotes to the number of levels for the reference image and similarly $h_q$ for the query image.\\
\textbf{Feature Augmentation.} Successful instance retrieval methods 
have many feature processing steps. Adopting the proposed pipeline
of \cite{JegouECCV12} and followed by others \cite{Gong14,Tolias13} we process the extracted 4096 dim features in the following way:
\L2 normalize $\rightarrow$ PCA dimensionality reduction $\rightarrow$
whitening $\rightarrow$ \L2 renormalization. Finally, we further use a signed
component wise power transform and raise each dimension of the feature
vector to the power of $2$. For all datasets in the PCA step we reduce
the dimensionality of the feature vector to 500. All the \L2
normalizations are applied to achieve unit length.


%% file: Retrieval_r.tex
The result of different retrieval methods applied to 5 datasets
are in
table \ref{tab:retrievalPerformance}. Spatial search is only used for the first three datasets which have samples in different scales and locations. For the other two datasets we used the same jittering as explained in Sec. \ref{sec:method}

\begin{table}
\scriptsize
\tabcolsep=0.1cm
\begin{center}
\begin{tabular}{l l l l l l l l l}
     \toprule
 &Dim&Oxford5k&Paris6k&Sculp6k&Holidays&UKBench\\
\midrule
\rowcolor[gray]{0.8}
BoB\cite{Arandjelovic11}
& N/A
& N/A 
& N/A
& \textbf{45.4}\cite{Arandjelovic11}
& N/A 
& N/A\\
BoW
&200k
& 36.4\cite{jegou12} 
& 46.0\cite{Philbin08}  
& 8.1\cite{Arandjelovic11} 
& 54.0\cite{Arandjelovic13} 
& 70.3\cite{jegou12}\\
     \rowcolor[gray]{0.8}
IFV\cite{perronnin10} 
&2k 
&41.8\cite{jegou12}  
& - 
& -
& 62.6\cite{jegou12} 
& 83.8\cite{jegou12} \\
VLAD\cite{Arandjelovic13}
& 32k
& 55.5 \cite{Arandjelovic13}
& -
& -
& 64.6\cite{Arandjelovic13}
& -\\
     \rowcolor[gray]{0.8}
CVLAD\cite{Zhao13} 
&64k
&47.8\cite{Zhao13}
&-
&-
&81.9\cite{Zhao13}
&89.3\cite{Zhao13}\\
HE+burst\cite{Jain11}
&64k
&64.5\cite{Tolias13}
&-
&-
&78.0\cite{Tolias13}
&-\\
\rowcolor[gray]{0.8}
AHE+burst\cite{Jain11}
&64k
&66.6\cite{Tolias13}
&-
&-
&79.4\cite{Tolias13}
&-\\

Fine vocab\cite{Mikulik10}
&64k
&74.2\cite{Mikulik10}
&74.9\cite{Mikulik10}
&-
&74.9\cite{Mikulik10}
&-\\
     \rowcolor[gray]{0.8}
ASMK*+MA\cite{Tolias13}
&64k
&80.4\cite{Tolias13}
&77.0\cite{Tolias13}
&-
&81.0\cite{Tolias13}
&-\\

ASMK+MA\cite{Tolias13}
&64k
&\textbf{81.7}\cite{Tolias13}
&78.2\cite{Tolias13}
&-
&82.2\cite{Tolias13}
&-\\
     \midrule
CNN
&4k
& 32.2 
& 49.5
& 24.1
& 64.2 
& 76.0\\
     \rowcolor[gray]{0.8}
CNN-ss
&32-120k
& 55.6
& 69.7 
& 31.1 
& 76.9 
& 86.9	\\
CNNaug-ss
&4-15k
&\textbf{68.0}
&\textbf{79.5}
&42.3
&\textbf{84.3}
&\textbf{91.1}\\
     \rowcolor[gray]{0.8}
     CNN+BOW\cite{Gong14}
     &2k
     &-
     &-
     &-
     &\textbf{80.2}
     &-\\
\bottomrule
\end{tabular}
\end{center}
\vspace{-0.5cm}
\caption{{\footnotesize \textbf{The result of object retrieval on 5 datasets.} All the
methods except the CNN have their representation trained on
datasets similar to those they report the results on.  
The spatial search result on Oxford5k,Paris6k and Sculpture6k, are reported for $h_r=4$  and $h_q=3$.
It can be seen that CNN features, when compared with low-memory footprint methods, produce consistent high results.
ASMK+MA \cite{Tolias13} and fine-vocab \cite{Mikulik10} use in order of million codebooks but with various tricks including binarization they reduce the memory foot print to 64k.}}
\label{tab:retrievalPerformance}
\vspace{-0.4cm}
\end{table}
It should be emphasized that we only reported the results on low memory footprint methods. 

%% file: Conclusion.tex
\section{Conclusion}
In this work, we used an off-the-shelf CNN representation,
\overfeat\hspace*{-3pt}, with simple classifiers to address
different recognition tasks. The learned CNN model was
originally optimized for the task of object classification
in ILSVRC 2013 dataset. Nevertheless, it showed itself
to be a strong competitor to the more sophisticated and
highly tuned state-of-the-art methods. The same trend
was observed for various recognition tasks and different
datasets which highlights the effectiveness and generality
of the learned representations. The experiments confirm and extend the
results reported in \cite{Donahue14}. We have also pointed to the
results from works which specifically \textit{optimize} the CNN representations for different tasks/datasets achieving even superior results. Thus, it can be concluded that
from now on, deep learning with CNN has to be considered as the
primary candidate in essentially any visual recognition task. 

%% file: Acknowledgment.tex
\paragraph{Acknowledgment.} We gratefully acknowledge the support of
NVIDIA Corporation with the donation of the Tesla K40 GPUs to this research. We further would like to thank Dr. Atsuto Maki, Dr. Pierre Sermanet, Dr. Ross Girshick, and Dr. Relja Arandjelovi\'c for their helpful comments.

%% file: CVPR14w.bbl
\begin{thebibliography}{10}\itemsep=-1pt

\bibitem{ilsvrc-2013}
Imagenet large scale visual recognition challenge 2013 (ilsvrc2013).
\newblock http://www.image-net.org/challenges/LSVRC/2013/.

\bibitem{Angelova13}
A.~Angelova and S.~Zhu.
\newblock Efficient object detection and segmentation for fine-grained
  recognition.
\newblock In {\em CVPR}, 2013.

\bibitem{Arandjelovic11}
R.~Arandjelovi\'c and A.~Zisserman.
\newblock Smooth object retrieval using a bag of boundaries.
\newblock In {\em ICCV}, 2011.

\bibitem{Arandjelovic13}
R.~Arandjelovi\'c and A.~Zisserman.
\newblock All about {VLAD}.
\newblock In {\em CVPR}, 2013.

\bibitem{Berg13}
T.~Berg and P.~N. Belhumeur.
\newblock Poof: Part-based one-vs.-one features for fine-grained
  categorization, face verification, and attribute estimation.
\newblock In {\em CVPR}, 2013.

\bibitem{Bourdev11}
L.~D. Bourdev, S.~Maji, and J.~Malik.
\newblock Describing people: A poselet-based approach to attribute
  classification.
\newblock In {\em ICCV}, 2011.

\bibitem{Chai11}
Y.~Chai, V.~S. Lempitsky, and A.~Zisserman.
\newblock Bicos: A bi-level co-segmentation method for image classification.
\newblock In {\em ICCV}, 2011.

\bibitem{Chen12}
Q.~Chen, Z.~Song, Y.~Hua, Z.~Huang, and S.~Yan.
\newblock Hierarchical matching with side information for image classification.
\newblock In {\em CVPR}, 2012.

\bibitem{Doersch13}
C.~Doersch, A.~Gupta, and A.~A. Efros.
\newblock Mid-level visual element discovery as discriminative mode seeking.
\newblock In {\em NIPS}, 2013.

\bibitem{Donahue14}
J.~Donahue, Y.~Jia, O.~Vinyals, J.~Hoffman, N.~Zhang, E.~Tzeng, and T.~Darrell.
\newblock Decaf: A deep convolutional activation feature for generic visual
  recognition.
\newblock In {\em ICML}, 2014.

\bibitem{Dong13}
J.~Dong, W.~Xia, Q.~Chen, J.~Feng, Z.~Huang, and S.~Yan.
\newblock Subcategory-aware object classification.
\newblock In {\em CVPR}, 2013.

\bibitem{voc-2012}
M.~Everingham, L.~Van~Gool, C.~K.~I. Williams, J.~Winn, and A.~Zisserman.
\newblock The {PASCAL} {V}isual {O}bject {C}lasses {C}hallenge 2012 {(VOC2012)}
  {R}esults.
\newblock
  http://www.pascal-network.org/challenges/VOC/voc2012/workshop/index.html.

\bibitem{Farhadi09}
A.~Farhadi, I.~Endres, D.~Hoiem, and D.~A. Forsyth.
\newblock Describing objects by their attributes.
\newblock In {\em CVPR}, 2009.

\bibitem{Fernando14}
B.~Fernando, E.~Fromont, and T.~Tuytelaars.
\newblock Mining mid-level features for image classification.
\newblock {\em International Journal of Computer Vision}, 2014.

\bibitem{Girshick13}
R.~B. Girshick, J.~Donahue, T.~Darrell, and J.~Malik.
\newblock Rich feature hierarchies for accurate object detection and semantic
  segmentation.
\newblock {\em arxiv:1311.2524 [cs.CV]}, 2013.

\bibitem{Gong14}
Y.~Gong, L.~Wang, R.~Guo, and S.~Lazebnik.
\newblock Multi-scale orderless pooling of deep convolutional activation
  features.
\newblock {\em CoRR}, 2014.

\bibitem{Jain11}
M.~Jain, H.~J{\'e}gou, and P.~Gros.
\newblock Asymmetric hamming embedding: taking the best of our bits for large
  scale image search.
\newblock In {\em ACM Multimedia}, pages 1441--1444, 2011.

\bibitem{JegouECCV12}
H.~J{\'e}gou and O.~Chum.
\newblock Negative evidences and co-occurences in image retrieval: The benefit
  of pca and whitening.
\newblock In {\em ECCV}, pages 774--787, 2012.

\bibitem{jegou08}
H.~J{\'e}gou, M.~Douze, and C.~Schmid.
\newblock Hamming embedding and weak geometric consistency for large scale
  image search.
\newblock In {\em ECCV}, 2008.

\bibitem{jegou12}
H.~J{\'e}gou, F.~Perronnin, M.~Douze, J.~S{\'a}nchez, P.~P{\'e}rez, and
  C.~Schmid.
\newblock Aggregating local image descriptors into compact codes.
\newblock {\em IEEE Transactions on Pattern Analysis and Machine Intelligence},
  34(9):1704--1716, 2012.

\bibitem{Juneja13}
M.~Juneja, A.~Vedaldi, C.~V. Jawahar, and A.~Zisserman.
\newblock Blocks that shout: Distinctive parts for scene classification.
\newblock In {\em CVPR}, 2013.

\bibitem{Krizhevsky12}
A.~Krizhevsky, I.~Sutskever, and G.~E. Hinton.
\newblock Imagenet classification with deep convolutional neural networks.
\newblock In {\em NIPS}, 2012.

\bibitem{Lampert14}
C.~H. Lampert, H.~Nickisch, and S.~Harmeling.
\newblock Attribute-based classification for zero-shot visual object
  categorization.
\newblock {\em IEEE Transactions on Pattern Analysis and Machine Intelligence},
  36(3), 2014.

\bibitem{Li10}
L.-J. Li, H.~Su, E.~P. Xing, and F.-F. Li.
\newblock Object bank: A high-level image representation for scene
  classification {\&} semantic feature sparsification.
\newblock In {\em NIPS}, 2010.

\bibitem{Li13}
Q.~Li, J.~Wu, and Z.~Tu.
\newblock Harvesting mid-level visual concepts from large-scale internet
  images.
\newblock In {\em CVPR}, 2013.

\bibitem{Mikulik10}
A.~Mikul\'{\i}k, M.~Perdoch, O.~Chum, and J.~Matas.
\newblock Learning a fine vocabulary.
\newblock In {\em ECCV}, pages 1--14, 2010.

\bibitem{Nilsback08}
M.-E. Nilsback and A.~Zisserman.
\newblock Automated flower classification over a large number of classes.
\newblock In {\em Proceedings of the Indian Conference on Computer Vision,
  Graphics and Image Processing}, Dec 2008.

\bibitem{nister06}
D.~Nist{\'e}r and H.~Stew{\'e}nius.
\newblock Scalable recognition with a vocabulary tree.
\newblock In {\em CVPR}, 2006.

\bibitem{Oquab13}
M.~Oquab, L.~Bottou, I.~Laptev, and J.~Sivic.
\newblock Learning and transferring mid-level image representations using
  convolutional neural networks.
\newblock Technical Report HAL-00911179, INRIA, 2013.

\bibitem{Pandey11}
M.~Pandey and S.~Lazebnik.
\newblock Scene recognition and weakly supervised object localization with
  deformable part-based models.
\newblock In {\em ICCV}, 2011.

\bibitem{Parizi12}
S.~N. Parizi, J.~G. Oberlin, and P.~F. Felzenszwalb.
\newblock Reconfigurable models for scene recognition.
\newblock In {\em CVPR}, 2012.

\bibitem{parkhi12}
O.~M. Parkhi, A.~Vedaldi, A.~Zisserman, and C.~V. Jawahar.
\newblock Cats and dogs.
\newblock In {\em CVPR}, 2012.

\bibitem{perronnin10}
F.~Perronnin, Y.~Liu, J.~S{\'a}nchez, and H.~Poirier.
\newblock Large-scale image retrieval with compressed fisher vectors.
\newblock In {\em CVPR}, 2010.

\bibitem{Philbin07}
J.~Philbin, O.~Chum, M.~Isard, J.~Sivic, and A.~Zisserman.
\newblock Object retrieval with large vocabularies and fast spatial matching.
\newblock In {\em CVPR}, 2007.

\bibitem{Philbin08}
J.~Philbin, O.~Chum, M.~Isard, J.~Sivic, and A.~Zisserman.
\newblock Lost in quantization: Improving particular object retrieval in large
  scale image databases.
\newblock In {\em CVPR}, 2008.

\bibitem{Quattoni09}
A.~Quattoni and A.~Torralba.
\newblock Recognizing indoor scenes.
\newblock In {\em CVPR}, 2009.

\bibitem{Rohrbach12}
M.~Rohrbach, S.~Amin, M.~Andriluka, and B.~Schiele.
\newblock A database for fine grained activity detection of cooking activities.
\newblock In {\em CVPR}, 2012.

\bibitem{Sermanet13}
P.~Sermanet, D.~Eigen, X.~Zhang, M.~Mathieu, R.~Fergus, and Y.~LeCun.
\newblock Overfeat: Integrated recognition, localization and detection using
  convolutional networks.
\newblock In {\em ICLR}, 2014.

\bibitem{Song11}
Z.~Song, Q.~Chen, Z.~Huang, Y.~Hua, and S.~Yan.
\newblock Contextualizing object detection and classification.
\newblock In {\em CVPR}, 2011.

\bibitem{Sun13}
J.~Sun and J.~Ponce.
\newblock Learning discriminative part detectors for image classification and
  cosegmentation.
\newblock In {\em ICCV}, 2013.

\bibitem{Taigman14}
Y.~Taigman, M.~Yang, M.~Ranzato, and L.~Wolf.
\newblock Deepface: Closing the gap to human-level performance in face
  verification.
\newblock In {\em CVPR}, 2014.

\bibitem{Tolias13}
G.~Tolias, Y.~S. Avrithis, and H.~J{\'e}gou.
\newblock To aggregate or not to aggregate: Selective match kernels for image
  search.
\newblock In {\em ICCV}, pages 1401--1408, 2013.

\bibitem{Toshev14}
A.~Toshev and C.~Szegedy.
\newblock Deeppose: Human pose estimation via deep neural networks.
\newblock In {\em CVPR}, 2014.

\bibitem{Tsagkatakis10}
G.~Tsagkatakis and A.~E. Savakis.
\newblock Sparse representations and distance learning for attribute based
  category recognition.
\newblock In {\em ECCV Workshops (1)}, pages 29--42, 2010.

\bibitem{Wah11}
C.~Wah, S.~Branson, P.~Welinder, P.~Perona, and S.~Belongie.
\newblock {The Caltech-UCSD Birds-200-2011 Dataset}.
\newblock Technical Report CNS-TR-2011-001, California Institute of Technology,
  2011.

\bibitem{Wang10}
Y.~Wang and G.~Mori.
\newblock A discriminative latent model of object classes and attributes.
\newblock In {\em ECCV}, 2010.

\bibitem{Yao11}
B.~Yao, A.~Khosla, and F.-F. Li.
\newblock Combining randomization and discrimination for fine-grained image
  categorization.
\newblock In {\em CVPR}, 2011.

\bibitem{Zeiler13}
M.~D. Zeiler and R.~Fergus.
\newblock Visualizing and understanding convolutional networks.
\newblock {\em CoRR}, abs/1311.2901, 2013.

\bibitem{Zhang12}
N.~Zhang, R.~Farrell, and T.~Darrell.
\newblock Pose pooling kernels for sub-category recognition.
\newblock In {\em CVPR}, 2012.

\bibitem{Zhang13}
N.~Zhang, R.~Farrell, F.~Iandola, and T.~Darrell.
\newblock Deformable part descriptors for fine-grained recognition and
  attribute prediction.
\newblock In {\em ICCV}, 2013.

\bibitem{Zhang14}
N.~Zhang, M.~Paluri, M.~Ranzato, T.~Darrell, and L.~Bourdev.
\newblock Panda: Pose aligned networks for deep attribute modeling.
\newblock In {\em CVPR}, 2014.

\bibitem{Zhao13}
W.-L. Zhao, H.~J{\'e}gou, G.~Gravier, et~al.
\newblock Oriented pooling for dense and non-dense rotation-invariant features.
\newblock In {\em BMVC}, 2013.

\end{thebibliography}


\begin{thebibliography}{10}\itemsep=-1pt

\bibitem{arandjelovic13}
R.~Arandjelovic and A.~Zisserman.
\newblock All about vlad.
\newblock In {\em CVPR}, pages 1578--1585, 2013.

\bibitem{Berg13}
T.~Berg and P.~N. Belhumeur.
\newblock Poof: Part-based one-vs.-one features for fine-grained
  categorization, face verification, and attribute estimation.
\newblock In {\em CVPR}, pages 955--962, 2013.

\bibitem{Chen12}
Q.~Chen, Z.~Song, Y.~Hua, Z.~Huang, and S.~Yan.
\newblock Hierarchical matching with side information for image classification.
\newblock In {\em CVPR}, pages 3426--3433, 2012.

\bibitem{Doersch13}
C.~Doersch, A.~Gupta, and A.~A. Efros.
\newblock Mid-level visual element discovery as discriminative mode seeking.
\newblock In {\em NIPS}, pages 494--502, 2013.

\bibitem{Dong13}
J.~Dong, W.~Xia, Q.~Chen, J.~Feng, Z.~Huang, and S.~Yan.
\newblock Subcategory-aware object classification.
\newblock In {\em CVPR}, pages 827--834, 2013.

\bibitem{jegou08}
H.~Jegou, M.~Douze, and C.~Schmid.
\newblock Hamming embedding and weak geometric consistency for large scale
  image search.
\newblock In {\em ECCV (1)}, pages 304--317, 2008.

\bibitem{jegou12}
H.~J{\'e}gou, F.~Perronnin, M.~Douze, J.~S{\'a}nchez, P.~P{\'e}rez, and
  C.~Schmid.
\newblock Aggregating local image descriptors into compact codes.
\newblock {\em IEEE Trans. Pattern Anal. Mach. Intell.}, 34(9), 2012.

\bibitem{Juneja13}
M.~Juneja, A.~Vedaldi, C.~V. Jawahar, and A.~Zisserman.
\newblock Blocks that shout: Distinctive parts for scene classification.
\newblock In {\em Proceedings of the {IEEE} Conf. on Computer Vision and
  Pattern Recognition ({CVPR})}, 2013.

\bibitem{Li10}
L.-J. Li, H.~Su, E.~P. Xing, and F.-F. Li.
\newblock Object bank: A high-level image representation for scene
  classification {\&} semantic feature sparsification.
\newblock In {\em NIPS}, pages 1378--1386, 2010.

\bibitem{Li13}
Q.~Li, J.~Wu, and Z.~Tu.
\newblock Harvesting mid-level visual concepts from large-scale internet
  images.
\newblock In {\em CVPR}, pages 851--858, 2013.

\bibitem{nister06}
D.~Nist{\'e}r and H.~Stew{\'e}nius.
\newblock Scalable recognition with a vocabulary tree.
\newblock In {\em CVPR (2)}, pages 2161--2168, 2006.

\bibitem{Pandey11}
M.~Pandey and S.~Lazebnik.
\newblock Scene recognition and weakly supervised object localization with
  deformable part-based models.
\newblock In {\em ICCV}, pages 1307--1314, 2011.

\bibitem{Parizi12}
S.~N. Parizi, J.~G. Oberlin, and P.~F. Felzenszwalb.
\newblock Reconfigurable models for scene recognition.
\newblock In {\em CVPR}, pages 2775--2782, 2012.

\bibitem{perronnin10}
F.~Perronnin, Y.~Liu, J.~S{\'a}nchez, and H.~Poirier.
\newblock Large-scale image retrieval with compressed fisher vectors.
\newblock In {\em CVPR}, pages 3384--3391, 2010.

\bibitem{Philbin07}
J.~Philbin, O.~Chum, M.~Isard, J.~Sivic, and A.~Zisserman.
\newblock Object retrieval with large vocabularies and fast spatial matching.
\newblock In {\em CVPR}, 2007.

\bibitem{Philbin08}
J.~Philbin, O.~Chum, M.~Isard, J.~Sivic, and A.~Zisserman.
\newblock Lost in quantization: Improving particular object retrieval in large
  scale image databases.
\newblock In {\em CVPR}, 2008.

\bibitem{Quattoni09}
A.~Quattoni and A.~Torralba.
\newblock Recognizing indoor scenes.
\newblock In {\em CVPR}, pages 413--420, 2009.

\bibitem{Song11}
Z.~Song, Q.~Chen, Z.~Huang, Y.~Hua, and S.~Yan.
\newblock Contextualizing object detection and classification.
\newblock In {\em CVPR}, pages 1585--1592, 2011.

\bibitem{Sun13}
J.~Sun and J.~Ponce.
\newblock Learning discriminative part detectors for image classification and
  cosegmentation.
\newblock In {\em ICCV}, pages 3400--3407, 2013.

\bibitem{Wah11}
C.~Wah, S.~Branson, P.~Welinder, P.~Perona, and S.~Belongie.
\newblock {The Caltech-UCSD Birds-200-2011 Dataset}.
\newblock Technical report, 2011.

\bibitem{Yao11}
B.~Yao, A.~Khosla, and F.-F. Li.
\newblock Combining randomization and discrimination for fine-grained image
  categorization.
\newblock In {\em CVPR}, pages 1577--1584, 2011.

\bibitem{Zhang12}
N.~Zhang, R.~Farrell, and T.~Darrell.
\newblock Pose pooling kernels for sub-category recognition.
\newblock In {\em CVPR}, pages 3665--3672, 2012.

\bibitem{Zhang13}
N.~Zhang, R.~Farrell, F.~Iandola, and T.~Darrell.
\newblock Deformable part descriptors for fine-grained recognition and
  attribute prediction.
\newblock In {\em ICCV}, pages 729--736, 2013.

\end{thebibliography}
